\documentclass{article}
\usepackage{ijcai18}

\usepackage{times}
\usepackage{graphicx}
\usepackage{xcolor}
\usepackage{soul}
\usepackage[utf8]{inputenc}
\usepackage[small]{caption}
\usepackage[linesnumbered, ruled, vlined]{algorithm2e}

\title{Multi-robot Path Planning in Well-formed Infrastructures: Prioritized Planning vs. Prioritized Wait Adjustment (Preliminary Results)}

\author{
Anton Andreychuk$^1$, 
Konstantin Yakovlev$^{2, 3}$
\\ 
$^1$ Peoples' Friendship University of Russia (RUDN University)\\
$^2$ Federal Research Center ``Computer Science and Control'' of Russian Academy of Sciences \\
$^3$ National Research University Higher School of Economics  \\
andreychuk@mail.com,
yakovlev@isa.ru
}

\begin{document}

\maketitle

\begin{abstract}
  We study the problem of planning collision-free paths for a group of homogeneous robots. We propose a novel approach for turning the paths that were planned egocentrically by the robots, e.g. without taking other robots' moves into account, into collision-free trajectories and evaluate it empirically. Suggested algorithm is much faster (up to one order of magnitude) than state-of-the-art but this comes at the price of notable drop-down of the solution cost.
  
\end{abstract}

\section{Introduction}

Multi-robot path planning is a challenging problem with the applications in transportation, logistics, video games etc. Commonly a discretized version of this problem is solved when the robots are confined to vertices of a graph capturing the connectivity of the shared workspace. Even in such case the problem is NP-hard to solve optimally minimizing the flowtime, e.g. the sum of robots traversal times, or the makespan, e.g. the time by which the last robot reaches its goal \cite{yu2016optimal}. Among the optimal algorithms the following can be referenced \cite{standley2010}, \cite{Wagner2011}, \cite{sharon2015}, \cite{yu2016optimal} etc. In general optimal solvers do no scale well to large problems (hundreds of agents) and can not handle them in a reasonable amount of time.

\begin{figure}[t]
    \centering
    \includegraphics[width=0.75\columnwidth]{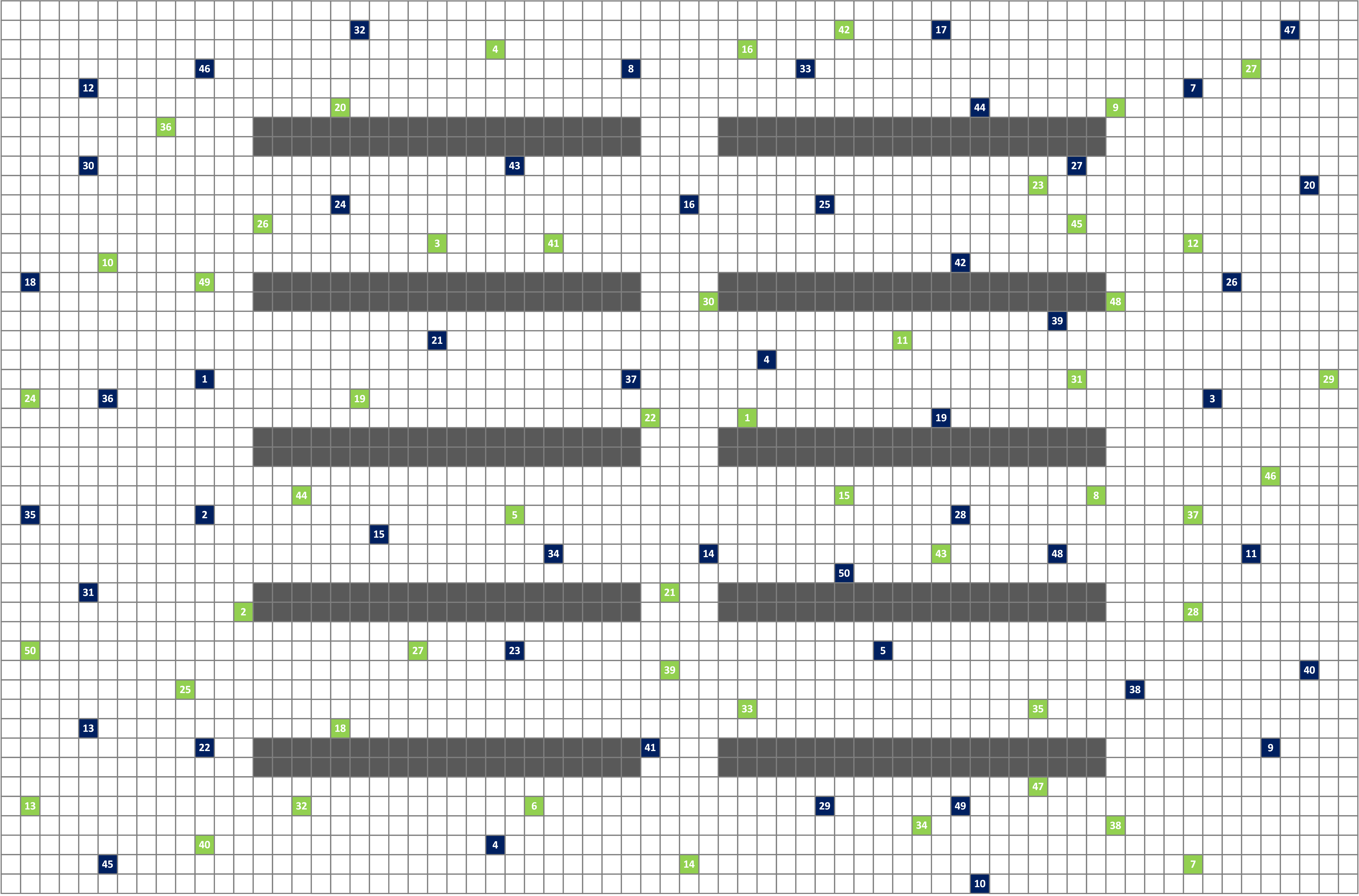}
    \caption{Well-formed infrastructure in the warehouse-like environment. Start locations are denoted in dark-blue, goal locations are light-green.}
    \label{fig1}
\end{figure}

One of the ways to increase the computational efficiency of multi-robot path finding is to use the prioritized approach \cite{erdmann1987}, when each robot is assigned a unique priority and then paths are planned sequentially one-by-one in accordance with the imposed ordering. The number of constraints a prioritized planner has to take into account is low as it treats all the previously planned trajectories as fixed and never backtracks. The downside is that one can not guarantee finding a solution in general. At the same time, it was shown in \cite{cap2015a} that slightly modified prioritized planner, e.g. the one explicitly avoiding start locations of other robots, is complete in so-called well-formed infrastructures (WFI). WFI is a multi-robot path finding instance that have all start and goal locations (endpoints) distributed in such way that any robot standing on the endpoint can not prevent other robots from finding their paths (see \figurename \ \ref{fig1}). WFI assumption rather often holds in practice, especially in logistics domain when the dedicated pick-up and drop-down locations have the adequate volume of free space between them.

Among other matters, WFIs have a nice property - any robot can occupy its start position for any period of time without a risk of being hit by other robots. This means a naive planner can be proposed that for a robot with the priority $k$ just waits for $t_1 + t_2 + ... + t_{k-1}$ timesteps, where $t_i$ is the time needed for the $i$-th robot to reach its goal, and then just finds a path without taking other robots moves into account, e.g. just finds a path in static environment. Obviously the quality of the overall solution will be very poor but the planner will produce solutions extremely fast. Taking inspiration from this idea, in this work we suggest a novel method to ``smartly'' adjust the duration of the wait actions for a robot following the path that was constructed without taking moving obstacles (other robots) into account. We evaluate the method empirically in simulated environments with 50-250 agents and compare it to state-of-the art prioritized planner and bounded sub-optimal conflict-based planner. Suggested algorithm is much faster (up to one order of magnitude) but, predictably, produces solutions of worse quality (flowtime/makespan is 1.2 - 2.8 times higher). Thus, the proposed method can be of particular value to time-critical multi-robot missions and applications, when one can sacrifice solution quality in order to get it as fast as possible. 

\begin{figure*}[h!]
    \centering
    \includegraphics[width=\textwidth]{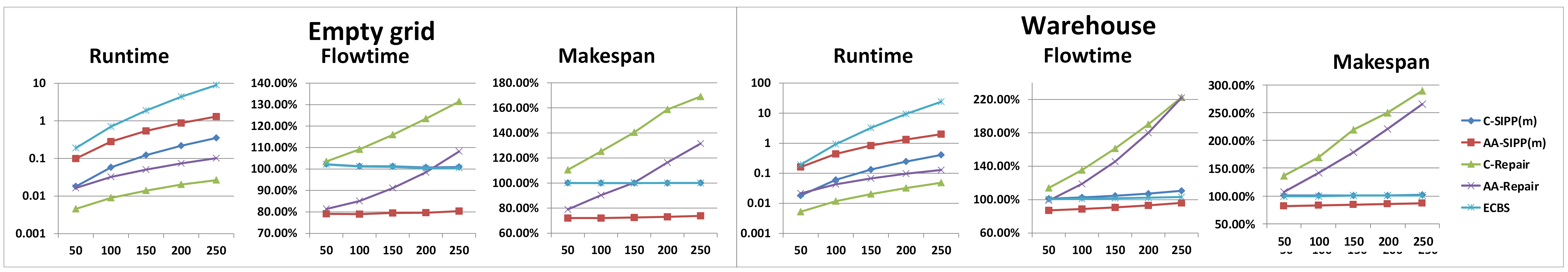}
    \caption{Results of the experimental evaluation.}
    \label{fig2}
\end{figure*}

\section{Method}
To check whether the conflict between the current robot, $i$, and some other high-priority robot exists we iterate through the segments $\pi_i$ and use the close-loop formula from \cite{guy2015} to detect collision. Consider now that the segment $[c_k, c_{k+1}]$ of  $\pi_i$ is a conflict-one, e.g. if the robot $i$ starts traversing the segment immediately after arriving to the cell $c_k$ collision occurs. Obviously one can avoid collision by waiting extra time so that the time robot leaves the start point is $t_{leave}=t_{arrive}(c_k) + t_{wait}(c_k)$. We propose incrementally increasing $t_{wait}$ by $\delta$ timepoints, where $\delta$ is the user-given parameter. Thus on the $j$-th step we set $t_{wait}$ to $\delta j$ and check whether the collision occurs. If yes -- the process repeats. Sooner or later the wait duration leading to no collisions will be found as other robots are contiguously moving towards their goals and thus essentially leave the impact zone.


The only problem that may arise now is that the computed wait time violates the constraints associated with the $c_k$, e.g. the robot $i$ will be hit by some high-priority robot while waiting in the cell $c_k$ for the desired amount of time. To formalize such constraints we use the notion of safe-interval (SI) as proposed in \cite{phillips2011}. SI is a ``contiguous period of time for a configuration, during which there is no collision and it is in collision one timestep prior and one timestep after the period''\footnote{Configuration is simply the grid cell in the considered case.}. Using approach from \cite{yakovlev2017aasipp} one can calculate all safe intervals for $c_k$. Knowing these SIs we need to check whether the computed $t_{leave}$ belongs to one of them. If it belongs to the same SI as $t_{arrive}$ then the robot can safely wait at $c_k$. If no, the robot must arrive at $c_k$ later - either by the time moment $t_{leave}$ in case $t_{leave}$ belongs to one of the safe-intervals of $c_k$, or by the beginning of SI which follows $t_{leave}$ in case $t_{leave}$ does not belong to any SI of $c_k$ (such SI always exists because $c_k$ is not a goal cell of some high-priority robot and all non-goal cells sooner or later become free). In both cases additional delay associated with $c_{k-1}$ is needed. Estimating the duration of this delay is based on the aligning the SIs of $c_{k-1}$ and $c_k$ in a proper way, i.e. in such a way that the move $[c_{k-1}, c_k]$ does not result in a collision. It might happen that this move is non-existent due to high-priority robots passing $c_{k-1}$. In this case additional wait time should be added to $c_{k-2}$ etc. In the worst case the wait location is shifted all the way back to $start_i$. Here the robot may wait as long as needed\footnote{Due to all robots explicitly avoid start locations of the others.} for the moves of $\pi_i$ up to $[c_k, c_{k+1}]$ to become collision-free.

After augmenting the path $\pi_i$ with the wait actions we fix the resultant trajectory and move on to processing $\pi_{i+1}$ until all $n$ paths are turned into conflict-free trajectories. This always happens in well-formed infrastructures when both start and goal locations of the robots are explicitly avoided by the path planner.

\section{Experimental evaluation}

Suggested approach was evaluated in simulated scenarios. We used A* for getting cardinal-only paths and Theta* for getting any-angle paths. The paths were augmented with the wait actions as described above and the resultant algorithms were dubbed ``C-Repair'' and ``AA-Repair''. We compared them against the prioritized planner from \cite{yakovlev2017aasipp} that uses the concept of safe-intervals and against bounded sub-optimal (w.r.t. flowtime) conflict-based planner ECBS \cite{barer2014}. ECBS handles only cardinal moves, while prioritized planner handles both cardinal and any-angle moves. We use labels ``C-SIPP(m)'' and ``AA-SIPP(m)'' to denote these two versions of the prioritized planner. 

$64 \times 64$ empty grid and $46 \times 70$ grid modeling the warehouse environment (see \figurename \ \ref{fig1}) were used for the evaluation. The number of agents varied from 50 to 250. 100 planning instances, all being well-formed infrastructures, per number of agents per environment were randomly generated.

Average runtime is depicted on \figurename \ \ref{fig2}. The scale is logarithmic. As one can see the proposed planner is extremely fast. It is one order of magnitude faster than the prioritized algorithm and two orders of magnitude faster than ECBS. The same figure shows averaged flowtime and makespan. These indicators are normalized using the conservative estimate of the lower bound of flowtime/makespan. A* cardinal-only paths that does not account for other agents were used to calculate this estimate, e.g. the lower bound of the flowtime is the sum of the lengths of such paths, makespan - is the length of the longest path. As one can note the quality of the solution degrades when the number of agents increases, especially for non-empty environments.



\section{Conclusion}
We have sketched a novel approach for multi-robot path planning tailored to solve special class of instances (well-formed infrastructures), which are commonly encountered in practice. It is based on modifying the time component of the trajectories without altering the spatial one. As a result it obtains solutions extremely fast, but their quality might be poor.

\subsubsection{Acknowledgments}
This work was partially supported by the Russian Science Foundation (Project No. 16-11-00048).
\bibliographystyle{named}
\bibliography{MAPF}

\end{document}